\def\year{2022}\relax
\documentclass[letterpaper]{article} 
\usepackage{aaai22}  
\usepackage{times}  
\usepackage{helvet}  
\usepackage{courier}  
\usepackage[hyphens]{url}  
\usepackage{graphicx} 
\usepackage{natbib}  
\usepackage{caption} 
\DeclareCaptionStyle{ruled}{labelfont=normalfont,labelsep=colon,strut=off} 
\def \R{{\rm I\!R}}
\def \x{\mathbf x}
\def \W{\mathbf W}
\def \w{\mathbf w}
\newcommand{\norm}[1]{\left\lVert#1\right\rVert^2 }

\usepackage{xcolor}
\usepackage{amssymb}
\usepackage{amsmath,amsfonts,bm}
\usepackage[mathscr]{euscript}

\renewcommand{\vec}[1]{\MakeLowercase{\mathbf{#1}}}
\renewcommand{\mod}[1]{\hspace{1pt}\text{mod}\hspace{1pt}#1}
\newcommand{\mat}[1]{\MakeUppercase{\mathbf{#1}}}
\newcommand{\tensor}[1]{\boldsymbol{\mathcal{{\MakeUppercase{#1}}}}}


\usepackage[ruled,vlined,noresetcount]{algorithm2e}
\SetKwComment{Comment}{/* }{ */}
\newlength\mylen

\newcommand{\nonl}{\renewcommand{\nl}{\let\nl\oldnl}}

\usepackage{amsthm}

\usepackage{thmtools,thm-restate}

\usepackage{booktabs}
\usepackage{float}
\usepackage{subcaption}
\title{Convolutional Neural Network Compression through \\ Generalized Kronecker Product Decomposition}
\author{
    Paper ID: 3803
}

\author {
    Marawan Gamal Abdel Hameed\textsuperscript{\rm 1,2}\thanks{Work done during an internship at Huawei Noah’s Ark Lab},
    Marzieh S. Tahaei\textsuperscript{\rm 1}\thanks{Corresponding Author}, 
    Ali Mosleh\textsuperscript{\rm 1},
    Vahid Partovi Nia\textsuperscript{\rm 1} 
    
}
\affiliations {
    \textsuperscript{\rm 1}Noah’s Ark Lab, Huawei Technologies Canada \\
    \textsuperscript{\rm 2}University of Waterloo \\
     marawan.abdelhameed@uwaterloo.ca,
     \{marzieh.tahaei,
     ali.mosleh,
     vahid.partovinia\}@huawei.com  \\

}
\begin{document}

\maketitle

\begin{abstract}

Modern Convolutional Neural Network (CNN) architectures, despite their superiority in solving various problems, are generally too large to be deployed on resource constrained edge devices.
In this paper, we reduce memory usage and floating-point operations required by convolutional layers in CNNs. We compress these layers by generalizing the Kronecker Product Decomposition to apply to multidimensional tensors, leading to the \emph{Generalized Kronecker Product Decomposition} (GKPD). Our approach yields a
plug-and-play module that can be used as a drop-in replacement for any convolutional layer. Experimental results for image classification on CIFAR-10 and ImageNet datasets using ResNet, MobileNetv2 and SeNet architectures substantiate the effectiveness of our proposed approach. We find that GKPD outperforms state-of-the-art decomposition methods including Tensor-Train and Tensor-Ring as well as other relevant compression methods such as pruning and knowledge distillation. 

\end{abstract}

\section{Introduction}


\noindent Convolutional Neural Networks (CNNs) have achieved  state-of-the-art performance on a wide range of computer vision tasks such as image classification \citep{he2016deep}, video recognition \citep{feichtenhofer2019slowfast} and object detection \citep{ren2015faster}. Despite achieving remarkably low generalization errors, modern CNN architectures are typically over-parameterized and consist of millions of parameters. As the size of state-of-the-art CNN architectures continues to grow, it becomes more challenging to deploy these models on resource constrained edge devices that are limited in both memory and energy.
%
Motivated by studies demonstrating that there is significant redundancy in CNN parameters \citep{Denil_Predicting_Parameters_in_Deep_Learning_2013}, model compression techniques such as pruning, quantization, tensor decomposition and knowledge distillation have emerged to address this problem.

\begin{figure}[t]
\begin{minipage}[c]{0.33\columnwidth}
\footnotesize
\vspace{1pt}
Original Tensor
\end{minipage}%
\begin{minipage}[c]{0.33\columnwidth}
\centering \footnotesize
Reconstructed Tensor
\end{minipage}%
\begin{minipage}[c]{0.33\columnwidth}
\footnotesize
\begin{flushright}
Reconstruction Error
\end{flushright}
\end{minipage}
\par\smallskip 
\centering
    \begin{subfigure}[b]{0.99\columnwidth}
        \includegraphics[width=\linewidth]{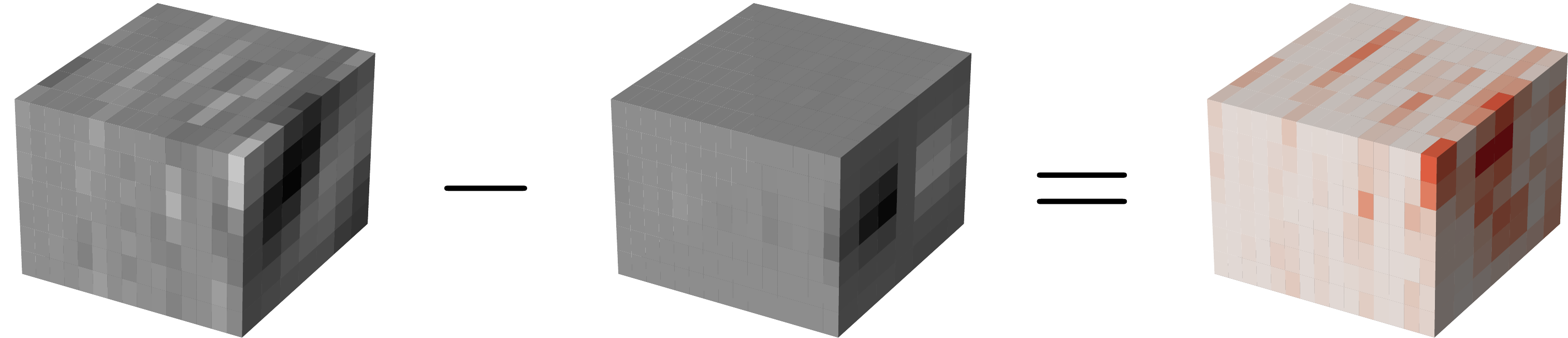} \vspace{-5mm}
    \caption{SVD (Tucker)}
    \end{subfigure}
    
    \begin{subfigure}[b]{0.99\columnwidth}
        \includegraphics[width=\linewidth]{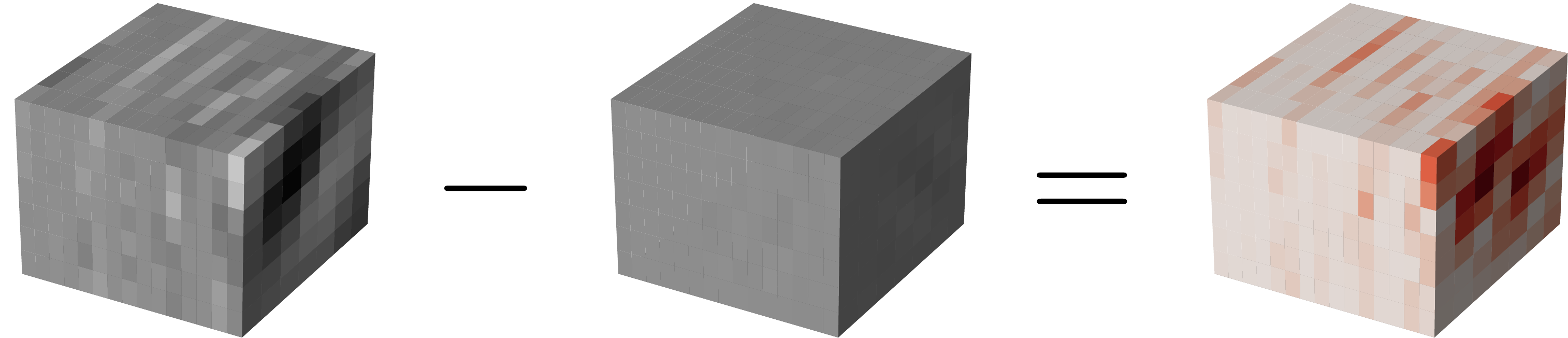} \vspace{-5mm}
    \caption{GKPD - 1}
    \end{subfigure}
    
    \begin{subfigure}[b]{0.99\columnwidth}
        \includegraphics[width=\linewidth]{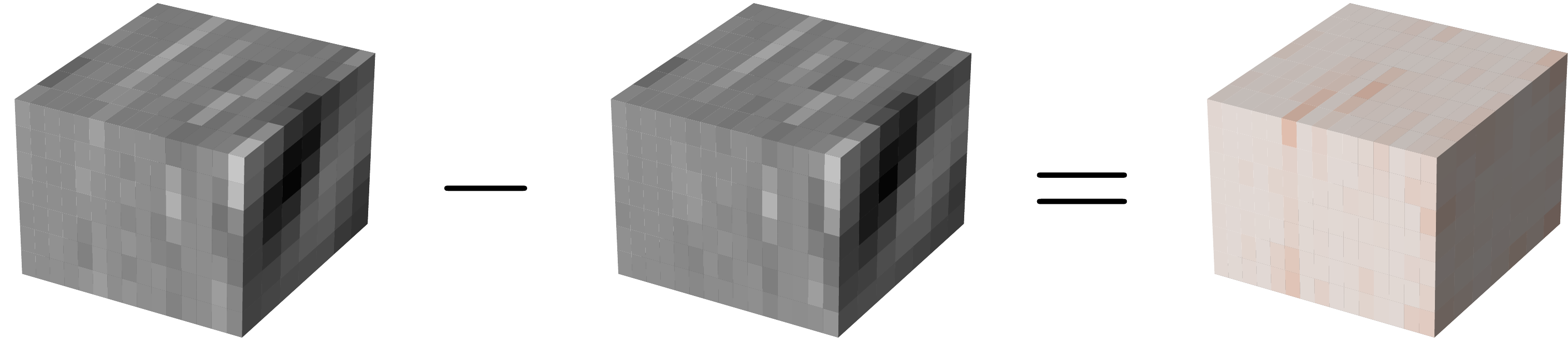} \vspace{-5mm}
    \caption{GKPD - 8}
    \end{subfigure} \vspace{-3mm}
\caption{
A compression rate of $2\times$ achieved for an arbitrary tensor from the first layer of ResNet18 using SVD (Tucker) in (a), and the proposed GKPD in (b) and (c). A larger summation, GKPD-8 achieves a lower reconstruction error in comparison with both a smaller summation, GKPD-1, as well as SVD (Tucker) decomposition.}
\label{fig:resnet18_first_layer_reconstruction}
\end{figure}

Decomposition methods have gained more attention in recent years as they can achieve higher compression rates in comparison to other approaches. Namely, Tucker \citep{Deok_Compression_of_Deep_Convolutional_Neural_Networks_for_Fast_and_Low_Power_Mobile_Applications_2016}, CP \citep{Lebedev_Speeding-up_Convolutional_Neural_Networks_Using_Fine-tuned_CP-Decomposition_2015}, Tensor-Train \citep{Garipov_Ultimate_Tensorization_2016} and Tensor-Ring \citep{Wang_Tensor_Ring_Nets_2018} decompositions have been widely studied for DNNs. However, these methods still suffer significant accuracy loss for computer vision tasks.


Kronecker Product Decomposition (KPD) is another decomposition method that has recently shown to be very effective when applied to RNNs \citep{thakker2019compressing}. KPD leads to model compression via replacing a large matrix with two smaller Kronecker factor matrices that best approximate the original matrix.
In this work, we generalize KPD to tensors, yielding the \emph{Generalized Kronecker Product Decomposition} (GKPD), and use it to decompose convolution tensors. GKPD involves finding the summation of Kronecker products between factor tensors that best approximates the original tensor. We provide a solution to this problem called the \emph{Multidimensional Nearest Kronecker Product Problem}. By formulating the convolution operation directly in terms of the Kronecker factors, we show that we can avoid reconstruction at runtime and thus obtain a significant reduction in memory footprints and floating-point operations (FLOPs). Once all convolution tensors in a pre-trained CNN have been replaced by their compressed counterparts, we retrain the network. If a pretrained network is not available, we show that we are still able to train our compressed network from a random initialization. Furthermore, we show that these randomly initialized networks retain universal approximation capability by building  on \cite{Hornik_UniversalMLP_1991} and \cite{Zhou_UniversalDeepCNN_2020}.

Applying GKPD to an arbitrary tensor leads to multiple possible decompositions, one for each configuration of Kronecker factors. As shown in Figure \ref{fig:resnet18_first_layer_reconstruction}, we find that for any given compression factor, choosing a decomposition that consists of a larger summation of smaller Kronecker factors (as opposed to a smaller summation of larger Kronecker factors) leads to a lower reconstruction error as well as improved model accuracy.

To summarize, the main contributions of this paper are:

\begin{itemize}
    \item Generalizing the  Kronecker  Product  Decomposition to multidimensional tensors
    \item Introducing the Multidimenesional Nearest Kronecker Product Problem and providing a solution
    \item Providing experimental results for image classification on CIFAR-10 and ImageNet using compressed ResNet \cite{he2016deep}, MobileNetv2 \cite{sandler2018mobilenetv2} and SeNet \cite{hu2018squeeze} architectures.
\end{itemize}




\section{Related Work on DNN Model Compression} \label{sec:related_work}



\textbf{Quantization} methods focus on reducing the precision of parameters and/or activations into lower-bit representations.
For example, the work in \citep{han2015deep} quantizes the parameter precision from 32 bits to 8 bits or lower. 
Model weights have been quantized even further into binary \citep{courbariaux2015binaryconnect,rastegari2016xnor,courbariaux2016binarized,hubara2017quantized}, and ternary \citep{li2016ternary,zhu2016trained} representations. 
In these methods, choosing between a uniform \citep{jacob2018quantization} or nonuniform \citep{han2015deep,tang2017train,zhang2018lq} quantization interval affects the compression rate and the acceleration.

\textbf{Pruning} methods began by exploring unstructured network weights and deactivating small weights through applying sparsity regularization to the weight parameters \citep{liu2015sparse,han2015deep,han2015learning} or  considering statistics information from layers to guide the parameter selections in ThiNet \citep{Luo_ThiNet_2017}. 
Unstructured pruning results in irregularities in the weight parameters which impact the expected acceleration rate of the pruned network. Hence, several works aim at zeroing out structured groups of the convolutional filters through group sparsity regularization \citep{zhou2016less,wen2016learning,alvarez2016learning}. Sparsity regularization has been combined with other forms of regularizers such as low-rank \citep{alvarez2017compression}, ordered weighted $\ell_1$ \citep{zhang2018learning}, and out-in-channel sparsity \citep{li2019oicsr} regularizers to further improve the pruning performance.

\textbf{Decomposition} methods factorize the original weight matrix or tensor into lightweight representations. This results in much fewer parameters and consequently fewer computations. Applying truncated singular value decomposition (SVD) to compress the weight matrix of fully-connected layers is one of the earliest works in this category \citep{denton2014exploiting}.
In the same vein, canonical polyadic (CP) decomposition of the kernel tensors was proposed in \citep{Lebedev_Speeding-up_Convolutional_Neural_Networks_Using_Fine-tuned_CP-Decomposition_2015}. This work uses nonlinear least squares to decompose the original convolution kernel into a set of rank-1 tensors (vectors).
An alternative tensor decomposition approach to convolution kernel compression is Tucker decomposition
\citep{tucker1963implications}.
Tucker decomposition has shown to provide more flexible interaction between the factor matrices through a core tensor. 
The idea of reshaping weights of fully-connected layers into high-dimensional tensors and representing them in Tensor-Train format \citep{oseledets2011tensor} was extended to CNNs in \citep{Garipov_Ultimate_Tensorization_2016}.
Tensor-Ring decomposition has also become another popular option to compress CNNs \citep{Wang_Tensor_Ring_Nets_2018}.  For multidimensional data completion with a same intermediate rank, TR can be far more expressive than TT \citep{wang2017efficient}.
Kronecker factorization was also used to replace the weight matrices and weight tensors
within fully-connected and convolution layers
\citep{Zhou_Exploiting_Local_Structures_with_the_Kronecker_Layer_in_Convolutional_Networks_2015}. This work however limited the representation to a single Kronecker product and trained the model with random initialization. As shown in Fig.\ref{fig:resnet18_first_layer_reconstruction} and in the next sections of this paper, summation can significantly improve the representation power of the network and thus leads to a performance increase. 

\textbf{Other model compression} forms can also be achieved through sharing convolutional weight matrices in a more structured manner as
ShaResNet \citep{boulch2018reducing} which reuses convolutional mappings within the same scale level
or FSNet \citep{Yang_FSNet_2020} which shares filter weights across spatial locations. 
NNs can also be compressed using Knowledge Distillation (KD) where a large (teacher) pre-trained network is used
to train a smaller (student) network \citep{Mirzadeh_Improved_Knowledge_Distillation_via_Teacher_Assistant_2020,Heo_Knowledge_Distillation_with_Adversarial_Samples_Supporting_Decision_2019}. Designing lightweight CNNs such as MobileNet \citep{sandler2018mobilenetv2} and SqueezeNet \citep{iandola2016squeezenet} is another form of model compression. 

    
    


\section{Preliminaries} \label{sec:prelim}

  
    Given matrices $\mat{a} \in \R^{m_1 \times n_1}$ and $\mat{b} \in \R^{m_2 \times n_2}$, their Kronecker product is the $m_1 m_2 \times n_1 n_2$ matrix 
    \begin{equation}
    \mat{a} \otimes \mat{b} \triangleq
    \begin{bmatrix}
      a_{1,1} \mat{b}  &  \dots  &  a_{1,n_1} \mat{b} \\
      \vdots          &  \ddots & \vdots \\
      a_{m_1,1} \mat{b}  &  \dots  &  a_{m_1,n_1} \mat{b} 
      \end{bmatrix}.  
      \label{eq:kron-matrix}
    \end{equation}
    As shown in \citet{van2000ubiquitous}, any matrix $\mat{w} \in \R^{m_1 m_2 \times n_1 n_2}$ can be decomposed into a sum of Kronecker products as
    \begin{equation}
      \mat{w} = \sum_{r=1}^R \mat{a}_r \otimes \mat{b}_r,
      \label{eq:kronecker_sum_matrix_reconstruction}
    \end{equation}
    where
    \begin{equation}
        R = \min(m_1n_1, m_2n_2)
    \end{equation}
    is the rank of a reshaped version of matrix $\mat{w}$. 
    We call this $R$ the \emph{Kronecker rank} of $\mat{w}$. Note that the Kronecker rank is not unique, and is dependent on the dimensions of factors $\mat{a}$ and $\mat{b}$. 
    
    To compress a given $\mat{w}$, we can represent it using a small number $\widehat{R} < R$ of Kronecker products that best approximate the original tensor. The factors are found by solving the Nearest Kronecker Product problem
    %
    \begin{equation}
      \underset{\{\mat{a}_r\}, \{\mat{b}_r\}}{\min} \left\| \mat{w} - \sum_{r=1}^{\widehat{R}}\mat{a}_r \otimes \mat{b}_r \right\|_F^2.
      \label{eq:nkp-matrices}
    \end{equation}
    As this approximation replaces a large matrix with a sequence of two smaller ones, memory consumption is reduced by a factor
    of 
    \begin{equation}
      \frac{ m_1 m_2 n_1 n_2 }{\widehat{R}(m_1 n_1 + m_2 n_2)}.
    \end{equation}
    Furthermore, if a matrix $\mat{w}$ is decomposed into its Kronecker factors then the projection $\mat{w}\Vec{x}$ can be performed without explicit reconstruction of $\mat{w}$. Instead, the factors can be used directly to perform the computation as a result of the following equivalency relationship:
    %
    %
    \begin{equation}
       \Vec{y} = (\mat{a} \otimes \mat{b})\Vec{x} 
       \equiv
       \mat{Y} = \mat{B} \mat{X} \mat{A}^\top, 
      \label{eq:kronecker-equivalency}
    \end{equation}
    where
 $\text{vec}(\mat{x})=\Vec{x}$, $\text{vec}(\mat{y}) = \Vec{y}$ and $\text{vec}(\cdot)$ vectorizes matrices $\mat{X} \in \R^{n_2 \times n_1}$ and $\mat{Y} \in \R^{m_2 \times m_1}$ by stacking their columns.
 


\section{Method} \label{sec:method}


In this section, we extend KPD to tensors yielding GKPD. First, we define the multidimensional Kronecker product, then we introduce the Multidimensional Nearest Kronecker Product problem and its solution. Finally, we describe our \emph{KroneckerConvolution} module that uses GKPD to compress convolution tensors and avoids reconstruction at runtime.

\subsection{Generalized Kronecker Product Decomposition}

\begin{figure}[t]
    \centering
    \includegraphics[width=0.8\columnwidth]{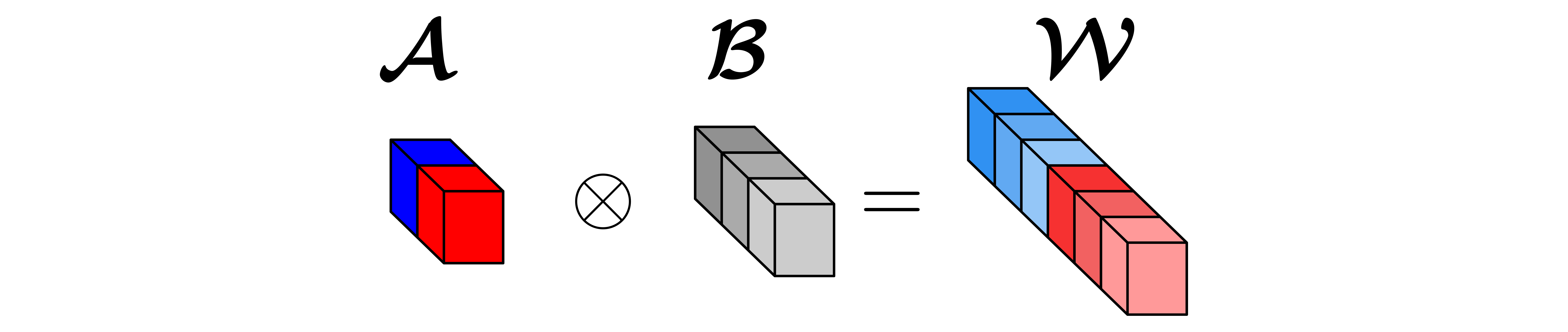}
\caption{Illustration of Kronecker Decomposition of a single convolution filter (with spatial dimensions equal to one for simplicity). 
}
    \label{fig:kpd_of_weight_tensor}
    \end{figure}

    We now turn to generalizing the Kronecker product to operate on tensors. Let $\tensor{a} \in \R^{a_1 \times a_2 \times \dots \times a_N}$ and $\tensor{b} \in \R^{b_1 \times b_2 \times \dots \times b_N}$ be two given tensors. Intuitively, tensor $(\tensor{a} \otimes \tensor{b}) \in \R^{a_1 b_1 \times a_2 b_2 \times \dots \times a_N b_N}$ is constructed by \emph{moving around} tensor $\tensor{B}$ in a non-overlapping fashion, and at each position scaling it by a corresponding element of $\tensor{A}$ as shown in Figure \ref{fig:kpd_of_weight_tensor}. Formally, the Multidimensional Kronecker product is defined as follows
    \begin{equation}
      (\tensor{A} \otimes \tensor{B})_{i_1, i_2, \dots, i_N} \triangleq \tensor{A}_{j_1, j_2, \dots, j_N} \tensor{B}_{k_1, k_2, \dots, k_N},
    \end{equation}
    where
    \begin{equation}
      j_n = \left \lfloor \frac{i_n}{b_n} \right\rfloor \text{and} \hspace{7pt}
      k_n = i_n\mod{b_n}
      \label{eq:kron-indexes}
    \end{equation}
    represent the integer quotient and the remainder term of $i_n$ with respect to divisor $b_n$, respectively.
    
    As with matrices, any multidimensional tensor $\tensor{w} \in \R^{w_1 \times w_2 \times \cdots \times w_N}$ can be decomposed into a sum of Kronecker products
    \begin{equation}
        \tensor{w} = \sum_{r=1}^{R} \tensor{a}_r \otimes  \tensor{b}_r, 
        \label{eq:kronecker_sum_tensor_reconstruction}
    \end{equation}
    where 
    \begin{equation}
        R = \min(a_1 a_2 \cdots a_N, b_1 b_2 \cdots b_N)
    \end{equation}
    denotes the Kronecker rank of tensor $\tensor{w}$. Thus, we can approximate $\tensor{w}$ using GKPD by solving the Multidimensional Nearest Kronecker Product problem
    %
    %
    \begin{equation}
      \underset{\{\tensor{a}_r\}, \{\tensor{b}_r\}}{\min} \left\| \tensor{w} - \sum_{r=1}^{\widehat{R}} \tensor{a}_r \otimes  \tensor{b}_r \right\|_F^2,
      \label{eq:nkp-tensor-problem}
    \end{equation}
    where $\widehat{R}$ $<$ $R$. For the case of matrices (2D tensors), \citet{Loan_Approximation_With_Kronecker_Products_1992} solved this problem using SVD. We extend their approach to the multidimensional setting. Our strategy will be to define rearrangement operators
    \begin{align*}
      \mat{R}_w  & : \R^{w_1 \times w_2 \times \dots \times w_N} \to \R^{a_1  a_2 \dots a_N \times b_1 b_2 \dots b_N} \\
      \vec{r}_a  &:  \R^{a_1 \times a_2 \times \dots \times a_N \to a_1 a_2 \dots a_N}\\
      \vec{r}_b &: \R^{b_1 \times b_2 \times  \dots \times b_N \to b_1 b_2 \dots b_N}
    \end{align*}
    %
    %
    %
    and solve
    \begin{equation}
      \underset{\{\tensor{a}_r\}, \{\tensor{b}_r\}}{\min} \left \| \mat{R}_w(\tensor{w}) - \sum_{r=1}^{\widehat{R}} \vec{r}_a(\tensor{a}_r)  \vec{r}_b(\tensor{b}_r) ^\top \right\|_F^2
      \label{eq:nkp-tensor-problem-reshaped}
    \end{equation}
    instead. By carefully defining the rearrangement operators, the sum of squares in \eqref{eq:nkp-tensor-problem-reshaped} is kept identical to that in \eqref{eq:nkp-tensor-problem}. The former corresponds to finding the best low-rank approximation which has a well known solution using SVD. We define the rearrangement operators as follows:
    \begin{align*}
          R_w(\tensor{w})_{i, :} &= \text{vec}(\text{unfold}(\tensor{w}, \Vec{d}_{\tensor{b}})_i) \\
      \vec{r}_a(\tensor{a}) &= \text{unfold}(\tensor{a},  \, \Vec{d}_{\tensor{I}_{\tensor{a}}}) \\
      \vec{r}_b(\tensor{b}) &= \text{vec}(\tensor{b})
    \end{align*}
        where
        \begin{equation*}
          \text{unfold}(\tensor{w},\, \Vec{d}): \R^{w_1 \times w_2 \times \dots \times w_N} \to \R^{N_p \times d_1 \times d_2 \dots d_N}
        \end{equation*} extracts $N_p$ non-overlapping patches of shape $\Vec{d}$ from tensor $\tensor{w}$, $\text{vec}(\cdot)$ flattens its input into a vector, tensor $\tensor{i}_{\tensor{a}}$ has the same number of dimensions as $\tensor{a}$ with each dimension equal to unity and $\Vec{d}_{\tensor{b}}$ is a vector describing the shape of tensor $\tensor{b}$.
        While the ordering of patch extraction and flattening is not important, it must remain consistent across the rearrangement operators.

      \begin{figure}[t]
        \centering
        \begin{subfigure}[t]{0.4\textwidth}
             \includegraphics[width=0.85\columnwidth,trim={0 600 0 0},clip]{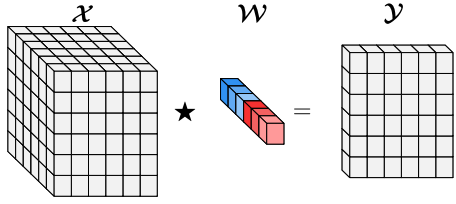}
            \caption{
            Conv2d} 
                   \bigskip
        \end{subfigure}
        \hfill
        \hfill
        \begin{subfigure}[t]{0.4\textwidth}
            \includegraphics[width=0.85\columnwidth,trim={0 200 0 0},clip]{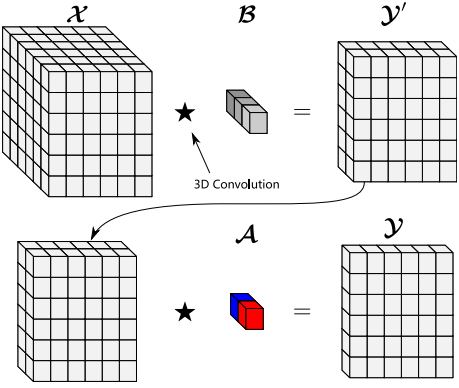}
            \caption{
            \vspace{-2mm}
            KroneckerConv2d}
        \end{subfigure}
        \caption{Illustration of the KroneckerConvolution operation. Although (a) and (b) result in identical outputs, the latter is more efficient in terms of memory and FLOPs.\vspace{-4mm}}
        \label{fig:kronecker-convolution}
    \end{figure}
    
\subsection{KroneckerConvolution Layer}

    The convolution operation in CNNs between a weight tensor $\tensor{W} \in \R^{F \times C \times K_w \times K_h}$ and an input $\tensor{x} \in \R^{C \times H \times W}$ is a multilinear map that can be described in scalar form as
    \begin{equation}
      \tensor{y}_{f,x,y} = \sum_{i=1}^{K_h} \sum_{j=1}^{K_w} \sum_{c=1}^{C} \tensor{W}_{f,c,i,j} \tensor{X}_{c,i+x,j+y}.
      \label{eq:scalar-form-conv}
    \end{equation}
    Assuming $\tensor{w}$ can be decomposed to KPD factors $\tensor{A} \in \R^{F_1 \times C_1 \times K_{w1} \times K_{h1}}$ and $ \tensor{B} \in \R^{F_2 \times C_2 \times K_{w2} \times K_{h2}}$, we can rewrite \eqref{eq:scalar-form-conv} as
    \begin{equation}
      \tensor{y}_{f,x,y} = \sum_{i=1}^{K_h} \sum_{j=1}^{K_w} \sum_{c=1}^{C} (\tensor{A} \otimes \tensor{B})_{f,c,i,j} \tensor{X}_{c,i+x,j+y}.
      \label{eq:scalar-form-conv-kpd1}
    \end{equation}
    Due to the structure of tensor $\tensor{A} \otimes \tensor{B}$, we do not need to explicitly reconstruct it to carry out the summation in \eqref{eq:scalar-form-conv-kpd1}. Instead, we can carry out the summation by \emph{directly} using elements of tensors $\tensor{a}$ and $\tensor{b}$ as shown in Lemma \ref{lem:kpd-equivalence}. This key insight leads to a large reduction in both memory and FLOPs. Effectively, this allows us to replace a large convolutional layer (with a large weight tensor) with two smaller ones, as we demonstrate in the rest of this section.
    
    \begin{restatable}{lem}{kpdequiv}
    \label{lem:kpd-equivalence}
    Suppose tensor $\tensor{w} \in \R^{w_1 \times w_2 \times \dots \times w_N}$ can be decomposed into KPD factors such that 
    $\tensor{w} = \tensor{a} \otimes \tensor{b}$.
    Then, the multilinear map involving $\tensor{w}$ can be written directly in terms of its factors $\tensor{A} \in \R^{a_1 \times a_2 \times \dots \times a_N}$  and $\tensor{B} \in \R^{b_1 \times b_2 \times \dots \times b_N}$ as follows
    \begin{multline*}
        \tensor{w}_{i_1, i_2, \dots, i_N} \tensor{x}_{i_1, i_2,\dots, i_N} =  \\
        \tensor{a}_{j_1, j_2, \dots, j_N} \tensor{b}_{k_1, k_2, \dots, k_N} \tensor{x}_{g(j_1, k_1), g(j_2, k_2), \dots, g(j_N, k_N)},
    \end{multline*}
    where $\tensor{x} \in \R^{d_1 \times d_2 \times \dots \times d_N}$ is an input tensor, $g(j_n, k_n) \triangleq j_n b_n + k_n$ is a re-indexing function; and $j_n,k_n$ are as defined in \eqref{eq:kron-indexes}. The equality also holds for any valid offsets to the input's indices 
    \begin{multline*}
        \tensor{w}_{i_1, i_2, \dots, i_N} \tensor{x}_{i_1 + o_1,\, i_2 + o_2,\, \dots,\, i_N + o_N} = 
        \tensor{a}_{j_1, j_2, \dots, j_N} \\ 
        \tensor{b}_{k_1, k_2, \dots, k_N} 
        \tensor{x}_{g(j_1, k_1) + o_1,\, g(j_2, k_2) + o_2,\, \dots,\, g(j_N, k_N) + o_N},
    \end{multline*}
    where $o_i \in \mathbb{N}$. 
    
    \end{restatable}
    \begin{proof}
        See Supplementary Material.
    \end{proof}
    
    \noindent Applying Lemma \ref{lem:kpd-equivalence} to the summation in \eqref{eq:scalar-form-conv-kpd1} yields
    \begin{multline*}
      \tensor{y}_{f,x,y} = \sum_{i_1, i_2} \sum_{j_1, j_2} \sum_{c_1, c_2} 
      \tensor{A}_{f_1, c_1, i_1, j_1} \tensor{B}_{f_2, c_2, i_2, j_2} \\
      \tensor{X}_{g(c_1,c_2), g(i_1, i_2)+x, g(j_1, j_2)+y},
    \end{multline*}
    where indices $i_1, j_1, c_1$ enumerate over elements in tensor $\tensor{a}$ and $i_2, j_2, c_2$ enumerate over elements in tensor $\tensor{b}$. Finally, we can separate the convolution operation into two steps by exchanging the order of summation as follows:
    \begin{multline}
        \tensor{y}_{f,x,y} = \sum_{i_1, j_1, c_1} \tensor{A}_{f_1, c_1, i_1, j_1}  \\ 
        \sum_{i_2, j_2, c_2}  \tensor{B}_{f_2, c_2, i_2, j_2} \tensor{X}_{g(c_1,c_2),g(i_1, i_2)+x,g(j_1, j_2)+y}.
        \label{eq:scalar-form-conv-kpd3}      
    \end{multline}
    The inner summation in \eqref{eq:scalar-form-conv-kpd3} corresponds to a 3D convolution and the outer summation corresponds to \emph{multiple} 2D convolutions, as visualized in Fig.~\ref{fig:kronecker-convolution} for the special case of $F=1$.
    
    
    \begin{algorithm}[t]
        \caption{Forward Pass}\label{alg:fw_pass}
        \KwIn{\\
            \Indp \Indp
            $\tensor{x} \in \R^{C \times W \times H}$ \\ 
            $\tensor{a}\in\R^{F_1 \times C_1 \times K_{h1} \times K_{w1}}$ \\
            $\tensor{b}\in\R^{F_2 \times C_2 \times K_{h2} \times K_{w2}}$ \\
            $\vec{s}_{\tensor{w}} \in \R^{4}$  // Stride of original convolution 
        }
        \KwOut{\\
        \Indp \Indp
        $\tensor{y} \in \R^{F \times W \times H}$ \\
        }
        \vspace{1mm}
        \vspace{1mm}
        $\tensor{x}' \gets \text{Unsqueeze}(\tensor{x}) \in \R^{1 \times C \times W \times H}$ \;
        \vspace{1mm}
        \tcc{3D Conv with stride of $(C_2, 1, 1)$}
        $\tensor{y}' \gets \text{Conv3d}(\tensor{b}, \tensor{x}') \in \R^{F_2 \times C_1 \times W \times H}$\;
        
        \vspace{1mm}
        \tcc{Batched 2D Conv with stride $\vec{s}_{\tensor{w}}$ and dilation $\vec{d}_{\tensor{b}} = \text{Shape}(\tensor{b})$. Note that we perform multiple 2D convolutions along the first dimension of size $F_2$ using the same weight kernel $\tensor{a}$}
        $\tensor{y}'' \gets \text{BatchConv2d}(\tensor{a}, \tensor{y}') \in \R^{F_2 \times F_1 \times W \times H}$\;
        
        \vspace{1mm}
        $\tensor{y} \gets \text{Reshape}(\tensor{y}'') \in \R^{F_1 F_2 \times W \times H}$\;
        
    \end{algorithm}

    
     Overall, \eqref{eq:scalar-form-conv-kpd3} can be carried out efficiently in tensor form using Algorithm \ref{alg:fw_pass}.
    Effectively, the input is collapsed in two stages instead of one as in the multidimensional convolution operation. Convolving a multi-channel input with a single filter in $\tensor{w}$ yields a scalar value at a particular output location.
    This is done by first scaling all elements in the corresponding multidimensional patch, then collapsing it by means of summation. Since tensor $\tensor{w}$ is comprised of multidimensional patches $\tensor{b}$ scaled by elements in $\tensor{a}$, we can equivalently collapse each \emph{sub-patch} in the input using tensor $\tensor{b}$ followed by a subsequent collapsing using tensor $\tensor{a}$ to obtain the same scalar value.

\subsection{Complexity of KroneckerConvolution}   
    The GKPD of a convolution layer is not unique. Different configurations of Kronecker factors will lead to different reductions in memory and number of operations. Namely, for a KroneckerConvolution layer using $\widehat{R}$ Kronecker products with factors $\tensor{a} \in \R^{F_1 \times C_1 \times K_{w1} \times K_{h1}}$ and $\tensor{b} \in \R^{F_2 \times C_2 \times K_{w2} \times K_{h2}}$ the memory reduction is
    
    \begin{equation}
      \frac{ F_1 C_1 K_{w1} K_{h1}  F_2 C_2 K_{w2} K_{h2}}{\widehat{R}(F_1 C_1 K_{w1} K_{h1} + F_2 C_2 K_{w2} K_{h2})},
    \end{equation}
    whereas the reduction in FLOPs is
    \begin{equation}
      \frac{F_1 C_1 K_{w1} K_{h1}  F_2 C_2 K_{w2} K_{h2}}{\widehat{R}(F_2 \cdot F_1 C_1 K_{w1} K_{h1} + C_1 \cdot F_2  C_2 K_{w2} K_{h2})}.
    \end{equation}
    %
    %
    %
    For the special case of using separable $3\times3$ filters, and $\widehat{R} = 1$ the reduction in FLOPs becomes
    \begin{equation}
        \frac{3F_1 C_2}{F_1  + C_2},
    \end{equation}
    implying that $F_1$ and $C_2$ should be sufficiently large in order to obtain a reduction in FLOPs. In contrast, memory reduction is unconditional in the KroneckerConvolution layer.


\subsection{Universal Approximation via Kronecker Products}



Universal approximation applied to shallow networks have been around for a long time \cite{Hornik_UniversalMLP_1991},\cite[pp 173--180]{Ripley_NNBook_1996} whilst such studies for deep networks are more recent \cite{Zhou_UniversalDeepCNN_2020}. In this section, we build off of these foundations to show that neural networks with weight tensors represented using low Kronecker rank summations of Kronecker products, remain universal approximators. For brevity, we refer to such networks as ``Kronecker networks''.

First, we show that a shallow Kronecker network is a universal approximator. For simplicity, this is shown only for one output. 
Then, we can generalize the resulting approximator via treating each output dimension separately.



    Consider a single layer neural network constructed using $n$ hidden units and an $L$-Lipschitz activation function $a(\cdot)$
    \begin{align*}
        \hat{f}_{\mat{w}}(x) & \triangleq \vec{w}_2 ^ \top a\left(\mat{W}\vec{x}\right) = \sum_{j=1}^n w_{2j} a\left(\w_{1j}^\top \x + w_{0j}\right),
    \end{align*}
    that is defined on a compacta $K$ in $\R^d$.  As shown in \cite{Hornik_UniversalMLP_1991}, such a network serves as a universal approximator, i.e., for a given positive number $\epsilon$ there exists an $n$ such that
    \begin{equation}
        \norm{f - \hat{f}_{\mat{W}}}_{2, \mu} \triangleq \int_K \left | f(\x) - \hat{f}_{\mat{w}}(\x) \right|^2  d\mu \\
        \leq \epsilon.
    \end{equation}
    Similarly, a shallow Kronecker network consisting of $n$ hidden units
    \begin{align}
        \hat{f}_{\mat{w}_{\widehat{R}}}(x) & \triangleq \vec{w}_2 ^ \top a\left(\mat{W}_{\widehat{R}}\vec{x}\right), \; 
        \mat{w}_{\widehat{R}} &= \sum_{r=1}^{\widehat{R}} \mat{a}_r \otimes \mat{b}_r, \label{eq:shallow-kronecker-network}
    \end{align}
    is comprised of a weight matrix $\mat{w}_{\widehat{R}}$ made of a summation of Kronecker products between factors $\mat{a}_r \in \R^{a_1 \times a_2}$ and $\mat{b}_r \in \R^{b_1 \times b_2}$. From \eqref{eq:shallow-kronecker-network}, we can see that any shallow neural network with $n$ hidden units can be represented exactly using a Kronecker network with a full Kronecker rank $R = \min(a_1 a_2, b_1 b_2)$. Thus, shallow Kronecker networks with full Kronecker rank also serve as universal approximators. 
    In Theorem \ref{theo:shallow} we show that a similar result holds for shallow Kronecker networks $f_{\mat{w}_{\widehat{R}}}$, with low Kronecker ranks $\widehat{R} < R$, provided that the $R-\widehat{R}$ smallest singular values of the reshaped matrix $R_w(\mat{W})$ of the approximating neural network $\hat{f}_{\mat{W}}$ are small enough.

\begin{restatable}{theo}{shallownet}
\label{theo:shallow}
 Any shallow Kronecker network with a low Kronecker rank $\widehat{R}$ and $n$ hidden units defined on a compacta $K\subset \R^d$ with $L$-Lipschitz activation is dense in the class of continuous functions $C(K)$ for a large enough $n$ given $$\sum\limits_{r=\hat R+1}^R \sigma^2_{r} < \epsilon (L \norm K  \norm{\w_2})^{-1},$$
 where $\sigma_r$ is the $r^\text{th}$ singular value of the reshaped version of the weight matrix  $R_w(\mat{W})$, in an approximating neural network $\hat{f}_{\mat{W}}$ with $n$ hidden units satisfying $\|f - \hat{f}_{\mat{W}}\|_{2,\mu}^2 < \epsilon$, for $f \in C(K)$.
\end{restatable}
\begin{proof}
See Supplementary Material.
\end{proof}


In Theorem \ref{theo:deep}, we extend the preceding result to deep convolutional neural networks, where each convolution tensor is represented using a summation of Kronecker products between factor tensors.

\begin{restatable}{theo}{deepconv}
\label{theo:deep}
Any deep Kronecker convolution network with Kronecker rank $\hat R_j$ in layer $j$ on compacta $K\subset \R^d$ with $L$-Lipschitz activation, is dense in the class of continuous functions $C(K)$ for a large enough number of layers $J$, given $$\prod_{j=1}^J 
\left(\sum_{r=\hat R_j+1}^{R_j}\sigma^2_{r, j}\right) < \epsilon(L^J\norm{\w_2} \norm{K})^{-1}, $$
where $\sigma_{r, j}$ is the $r^\text{th}$ singular value of the matrix $R_w(\tensor{w}^{j})$ of the reshaped weight tensor in the $j^\text{th}$ layer of an approximating convolutional neural network.
\end{restatable}
\begin{proof}
See Supplementary Material.
\end{proof}
The result is achieved by extending the recent universal approximation bound \cite{Zhou_UniversalDeepCNN_2020} for the GKPD networks. One can derive the convergence rates using \cite[Theorem 2]{Zhou_UniversalDeepCNN_2020} as well. 
These results assure that the performance degradation of Kronecker networks is small, in comparison to uncompressed networks, for an appropriate choice of Kronecker rank $\widehat{R}$.

\subsection{Configuration Setting}

As GKPD provides us with a set of possible decompositions for each layer in a network, a selection strategy is needed. For a given compression rate, there is a trade-off between using a larger number of terms $\widehat{R}$ in the GKPD summation \eqref{eq:nkp-tensor-problem} together with a more compressed configuration and a smaller $\widehat{R}$ with a less compressed configuration. To guide our search, we select the decomposition that best approximates the original uncompressed tensor obtained from a pretrained network. This means different layers in a network will be approximated by a different number of Kronecker products. Before searching for the best decomposition, we limit our search space to configurations that satisfy a desired reduction in FLOPs. Unless otherwise stated all GKPD experiments use this approach.

\section{Experiments}

 To validate our method, we provide model compression experimental results for image classification tasks using a variety of popular CNN architectures such as ResNet \citep{he2016deep}, and SEResNet which benefits from the squeeze-and-excitation blocks \citep{hu2018squeeze}.
 We also choose to apply our compression method on MobileNetV2 \citep{sandler2018mobilenetv2} as a model that is optimized for efficient inference on embedded vision applications through depthwise separable convolutions and inverted residual blocks. We provide implementation details in the Supplementary Material.


Table~\ref{tbl:cifar10_top1_different_models} shows the top-1 accuracy on the CIFAR-10 \cite{krizhevsky2012cifar10} dataset using compressed ResNet18 and SEResNet50. For each architecture, the compressed models obtained using the proposed GKPD are named with the ``Kronecker'' prefix added to the original model's name. The configuration of each compressed model is selected such that the number of parameters is similar to MobileNetV2.
We observe that for ResNet18 and SEResNet50, the number of parameters and FLOPs can be highly lowered at the expense of a small decrease in accuracy. Specifically, KroneckerResNet18 achieves a compression of 5$\times$ and a 4.7$\times$ reduction in FLOPs with only 0.08\% drop in accuracy. KroneckerSEResNet50 obtains a compression rate of 9.3$\times$ and a 9.7$\times$ reduction in FLOPs with only 0.7\% drop in accuracy.

Moreover, we see that applying the proposed GKPD method on higher-capacity architectures such as ResNet18 and SEResNet50 can lead to higher accuracy than a hand-crafted efficient network such as MobileNetV2. Specifically, with the same number of parameters as that of MobileNetV2, we achieve a compressed ResNet18 (KroneckerResNet18) and a compressed SEResNet50 (KroneckerSEResNet50) with 0.80\% and 0.27\%  higher accuracy than MobileNetV2.

Table \ref{tbl:cifar10_top1_extreme_compression} shows the performance of GKPD when used to achieve extreme compression rates. The same baseline architectures are compressed using different configurations.  
We also use GKPD to compress the already efficient MobileNetV2. When targeting very small models (e.g., 0.29M parameters) compressing MobileNetV2 with a compression factor of 7.9$\times$ outperforms extreme compression of SEResNet50 with a compression factor of 73.79$\times$.

In the following subsections, we present comparative assessments using different model compression methods. 
   
       \begin{table}[t]
        \resizebox{\columnwidth}{!}{
    \begin{tabular}{@{}lccc@{}}
        \toprule
        Model &  Params (M)  & FLOPs (M) & Accuracy(\%)  \\
        \midrule
        MobileNetV2 (Baseline) & 2.30 & 96 & 94.18  \\ 
        \midrule
        ResNet18 (Baseline)   & 11.17  & 557 & 95.05 \\
        KroneckerResNet18  & 2.2  & 117 & 94.97 \\
        \midrule
        SEResNet50 (Baseline) & 21.40 & 1163  & 95.15 \\
        KroneckerSeResNet50 & 2.30 & 120 & 94.45\\
    \bottomrule
    \end{tabular}
        }
        \caption{Top-1 accuracy measured on CIFAR-10 for the baseline models MobileNetV2, ResNet18 and SEResNet as well their compressed versions using GKPD. The number of parameters in compressed models are approximately matched with that of MobileNetV2.
        }
        \label{tbl:cifar10_top1_different_models}
    \end{table}
     \begin{table}[t]
        \resizebox{\columnwidth}{!}{
    \begin{tabular}{@{}lccc@{}}
        \toprule
        Model &  Params (M) & Compression & Accuracy(\%)  \\ \midrule
        KroneckerResNet18  & 0.48  & 23.27$\times$  & 92.62 \\
        \midrule
        KroneckerSeResNet50 & 0.93 & 23.01$\times$ & 93.66 \\
        KroneckerSeResNet50 & 0.29 & 73.79$\times$ & 91.85 \\
         \midrule
        KroneckerMobileNetV2 &  0.73 & 3.15$\times$  & 93.80\\ 
        KroneckerMobileNetV2 &  0.29 & 7.90$\times$ & 93.01 \\
        KroneckerMobileNetV2 & 0.18 & 12.78$\times$ & 91.48 \\
    \bottomrule
    \end{tabular}
        }
        \caption{Top-1 accuracy measured on CIFAR-10 highly compressed  ResNet18 \citep{he2016deep}, MobileNetV2 \citep{sandler2018mobilenetv2} and SEResNet \citep{hu2018squeeze}.   
    }
        \label{tbl:cifar10_top1_extreme_compression}
    \end{table}
    
\subsection{Comparison with Decomposition-based Methods}
    
In this section, we compare GKPD to other tensor decomposition compression methods. 
We use a classification model pretrained on CIFAR-10 and apply model compression methods based on Tucker \citep{Deok_Compression_of_Deep_Convolutional_Neural_Networks_for_Fast_and_Low_Power_Mobile_Applications_2016}, Tensor-Train  \citep{Garipov_Ultimate_Tensorization_2016}, and Tensor-Ring \citep{Wang_Tensor_Ring_Nets_2018}, along with our proposed GKPD method. We choose ResNet32 architecture in this set of experiments since it has been reported to be effectively compressed using Tensor-Ring in \citep{Wang_Tensor_Ring_Nets_2018}.
    
The model compression results obtained using different decomposition methods aiming for a 5$\times$ compression rate are shown in Table~\ref{tbl:cifar10_top1_decompostion}.  
As this table suggests, GKPD outperforms all other decomposition methods for a similar compression factor.
We attribute the performance of GKPD to its higher representation power. This is reflected in its ability to better reconstruct weight tensors in a pretrained network in comparison to other decomposition methods.
Refer to Supplementary Material for a comparative assessment of reconstruction errors for different layers of the ResNet architecture. 

    \begin{table}[t]
        \resizebox{\columnwidth}{!}{
        \begin{tabular}{@{}lccc@{}}
            \toprule
            Model & Params (M) & Compression & Accuracy (\%) \\
            \midrule
            Resnet32 & 0.46 & 1$\times$ & 92.55   \\
            \midrule
            TuckerResNet32 & 0.09 & 5$\times$ & 87.7 \\
            TensorTrainResNet32  & 0.096 & 4.8$\times$ & 88.3 \\
            TensorRingResNet32 & 0.09 & 5$\times$ & 90.6 \\
            KroneckerResNet32  & 0.09 & 5$\times$  & \textbf{91.52} \\
            \bottomrule
        \end{tabular}
        }
        \caption{Top-1 Accuracy on CIFAR-10 of compressed ResNet32 models using various decomposition approaches.
        }
        \label{tbl:cifar10_top1_decompostion}
    \end{table}
    \begin{table}[t]
    \resizebox{\columnwidth}{!}{
    \begin{tabular}{@{}lccc@{}}
        \toprule
        Model  & Params (M) & Compression & Accuracy (\%) \\
        \midrule
        ResNet26 & 0.37 & 1$\times$ & 92.94 \\
        \midrule
        \citet{Mirzadeh_Improved_Knowledge_Distillation_via_Teacher_Assistant_2020} & 0.17 & 2.13$\times$ & 91.23 \\
     \citet{Heo_Knowledge_Distillation_with_Adversarial_Samples_Supporting_Decision_2019} & 0.17 & 2.13$\times$ & 90.34 \\
        KroneckerResNet26 & 0.14 & 2.69$\times$ & \textbf{93.16} \\
        \midrule
      \citet{Mirzadeh_Improved_Knowledge_Distillation_via_Teacher_Assistant_2020} & 0.075 & 4.88$\times$ & 88.0 \\
      \citet{Heo_Knowledge_Distillation_with_Adversarial_Samples_Supporting_Decision_2019} & 0.075 & 4.88$\times$ & 87.32 \\
        KroneckerResNet26 & 0.069 & 5.29$\times$ & \textbf{91.28} \\
        \bottomrule
    \end{tabular}
    }
    \caption{Top-1 accuracy measured on CIFAR-10 for the baseline model ResNet26 and its compressed versions obtained using the KD-based methods;  \citep{Mirzadeh_Improved_Knowledge_Distillation_via_Teacher_Assistant_2020}, \citep{Heo_Knowledge_Distillation_with_Adversarial_Samples_Supporting_Decision_2019}, and the proposed GKPD method. 
    }
    \label{tbl:cifar10_top1_distillation}
\end{table}

    \begin{table}[t]
        \resizebox{\columnwidth}{!}{
        \begin{tabular}{@{}lccc@{}}
            \toprule
            Model  & Params (M) & Compression & Accuracy (\%) \\
            \midrule
            ResNet50 & 25.6 & 1$\times$ & 75.99 \\
            \midrule
            FSNet & 13.9 & 2.0$\times$ & 73.11 \\ 
            ThiNet  & 12.38 & 2.0$\times$ & 71.01 \\
            KroneckerResNet50 & 12.0 & 2.13$\times$ & \textbf{73.95}\\
            \bottomrule
        \end{tabular}
        }
        \caption{Top-1 accuracy measured on ImageNet for the baseline model ResNet50 and its compressed versions obtained using ThiNet \cite{Luo_ThiNet_2017}, FSNet \cite{Yang_FSNet_2020}, and the proposed GKPD method.
        }
        \label{tbl:imagenet_top1}
    \end{table}
    \begin{table}[t]
        \resizebox{\columnwidth}{!}{
     \begin{tabular}{@{}lccc@{}}
        \toprule
        Model & Params (M)  & FLOPs (M) & Accuracy(\%)  \\ \midrule
        ResNet18 & 11.17 &  0.58 & 95.05 \\
        \midrule
        KroneckerResNet18 $(\widehat{R}=4)$ & 1.41  & 0.17 & 92.96 \\
        KroneckerResNet18 $(\widehat{R}=8)$ & 1.42  & 0.16 & 93.74 \\
        KroneckerResNet18 $(\widehat{R}=16)$ & 1.44 & 0.26 & 94.30 \\
        KroneckerResNet18 $(\widehat{R}=32)$ & 1.51 & 0.32& 94.58 \\
        \bottomrule
    \end{tabular}
        }
        %
        \caption{Top-1 image classification accuracy of compressed ResNet18 on CIFAR-10, where $\widehat{R}$ denotes the number of Kronecker products used in the GKPD of each layer.
        }
        \label{tbl:cifar-10-ablation}
    \end{table}

\subsection{Comparison with other Compression Methods}

We compare our proposed model compression method with two state-of-the-art KD-based compression methods; \citep{Mirzadeh_Improved_Knowledge_Distillation_via_Teacher_Assistant_2020} and \citep{Heo_Knowledge_Distillation_with_Adversarial_Samples_Supporting_Decision_2019}. These methods are known to be very effective on relatively smaller networks such as ResNet26. Thus, we perform our compression method on ResNet26 architecture in these experiments. Table \ref{tbl:cifar10_top1_distillation} presents the top-1 accuracy obtained for different compressed models with two different compression rates. As this table suggests, the proposed method results in greater than 2$\%$ and 3.7$\%$ improvements in top-1 accuracy once we aim for compression rates of $\sim$2$\times$ and $\sim$5$\times$, respectively, compared to using the KD-based model compression methods.

\subsection{Model Compression with Random Initialization}
To study the effect of replacing weight tensors in neural networks with a summation of Kronecker products, we conduct experiments using randomly initialized Kronecker factors as opposed to performing GKPD on a pretrained network. By replacing all weight tensors in a predefined network architecture with a randomly initialized summation of Kronecker products, we obtain a compressed model.
To this end, we run assessments on a higher capacity architecture i.e, ResNet50 on a larger scale dataset i.e, ImageNet \cite{krizhevsky2012imagenet}.     
Table~\ref{tbl:imagenet_top1} lists the top-1 accuracy for ResNet50 baseline and its compressed variation. We achieve a compression rate of 2.13$\times$ with a 2$\%$ accuracy drop compared to the baseline model.

We also perform model compression using two state-of-the-art model compression methods; ThiNet \cite{Luo_ThiNet_2017} and FSNet \cite{Yang_FSNet_2020}. ThiNet and FSNet are based on pruning and filter sharing techniques, respectively. They both reportedly, lead to a good accuracy on large datasets. Table~\ref{tbl:imagenet_top1} also lists the top-1 accuracy for ResNet50 compressed using these two methods. As the table shows, our proposed method outperforms the other two techniques for a $\sim$2$\times$ compression rate. 
Note that the performance obtained using our method is based on a random initialization, while the compression achieved with ThiNet benefits from a pretrained model. These results indicate that the proposed GKPD can lead to a high performance even if a pretrained model is not available.

\subsection{Experimental Analysis of Kronecker Rank}

Using a higher Kronecker rank $\widehat{R}$ can increase the representation power of a network. This is reflected by the ability of GKPD to better reconstruct weight tensors using a larger number of Kronecker products in \eqref{eq:nkp-tensor-problem}. In Table~\ref{tbl:cifar-10-ablation} we study the effect of using a larger $\widehat{R}$ in Kronecker networks while keeping the overall number of parameters constant. We find that using a larger $\widehat{R}$ does indeed improve performance.

    
\section{Conclusion}      
In this paper we propose GKPD, a generalization of Kronecker Product Decomposition to multidimensional tensors for compression of deep CNNs. In the proposed GKPD, we extend the Nearest Kronecker Product problem to the multidimensional setting and use it for optimal initialization from a baseline model. 
We show that for a fixed number of parameters, using a summation of Kronecker products can significantly increase the representation power in comparison to a single Kronecker product. We use our approach to compress a variety of CNN architectures and show the superiority of GKPD to some state-of-the-art compression methods.
GKPD can be combined with other compression methods like quantization and knowledge distillation to further improve the compression-accuracy trade-off. Designing new architectures that can benefit most from Kronecker product representation is an area for future work.

\section*{Acknowledgments}
The authors thank Ali Ghodsi and Guillaume Rabusseau for useful discussions and suggestions.

  \bibliography{bibfile}

\appendix
\section*{Appendix}

\subsection{Implementation Details}
All experiments on CIFAR-10 are run for 200 epochs using Stochastic Gradient Descent (SGD). We use a batch size of 128, weight decay of 0.0001, momentum of 0.1 and an initial learning rate of 0.1 that is subsequently reduced by a factor of 10 at epochs 100 and 150. Similarly, experiments on ImageNet are run for 100 epochs using SGD with a batch size of 256, weight decay of 0.0001, momentum of 0.1 and an initial learning rate of 0.1 that is subsequently reduced by a factor of 10 at epochs 30, 60 and 90. Eight NVIDIA Tesla V100 SXM2 32 GB GPUs were used to run all of our experiments.


\subsection{Theorem Proofs}

\kpdequiv*
\begin{proof}
    By definition the terms in tensor $\tensor{w}$ are given by
    \begin{equation}
        \tensor{w}_{i_1, i_2, \dots, i_N} \triangleq \tensor{A}_{j_1, j_2, \dots, j_N} \tensor{B}_{k_1, k_2, \dots, k_N}
        \label{eq:lemma-kpd}
    \end{equation}
    where
    \begin{equation*}
        j_n = \left \lfloor \frac{i_n}{b_n} \right\rfloor, \hspace{7pt}
        k_n = i_n\mod{b_n}
    \end{equation*}
    
    Since $j_n$ and $k_n$ decompose $i_n$ into an integer quotient and a remainder term (with respect to divisor $b_n$), it follows that
    \begin{equation}
        g(j_n, k_n) \triangleq j_n b_n + k_n = i_n
        \label{eq:lemma-quotient-and-remainder}
    \end{equation}
    Therefore, 
    \begin{multline}
        \tensor{x}_{i_1 + o_1,\, i_2 + o_2,\, \dots, i_N + o_N} = \\ \tensor{x}_{g(j_1, k_1)+ o_1,\, g(j_2, k_2) + o_2,\, \dots,\, g(j_N, k_N) + o_N}
        \label{eq:lemma-reindexing}
    \end{multline}
    Combining \eqref{eq:lemma-kpd} and \eqref{eq:lemma-reindexing} completes the proof.
  \end{proof}
  
\shallownet*
\begin{proof}
We drop the bias term for the simplicity of notation. We need to bound 
    \begin{eqnarray}
    && \norm {f - \hat{f}_{\W_{\widehat{R}}}}_{2, \mu} = \int_K (f(\x) - \hat{ f}_{\W_{\widehat{R}}} (\x) )^2 d\mu  \nonumber \\ 
    &=& \int_K (f(\x) - \hat f_{\W}(\x))^2 d\mu \label{eq:first} \\ 
    & + &\int_K (\hat f_{\W}(\x) - \hat f_{\W_{\widehat{R}}}(\x))^2 d\mu \label{eq:second}\\
    &+& 2 \int_K (f(\x) - \hat{ f}_{\W}(\x)) \nonumber \\
    & & \quad\qquad (\hat f_{\W}(\x) - \hat{f}_{\W_{\widehat{R}}}(\x)) d\mu \label{eq:third}
    \end{eqnarray}
The full rank version $\hat f_\W$ is dense in $C(K)$ according to \cite{Hornik_UniversalMLP_1991}, therefore \eqref{eq:first} is bounded by $\epsilon$. It is only required to show that \eqref{eq:second} is also bounded
\begin{eqnarray*}
&& \int_K (\hat f_{\W}(\x) - \hat f_{\W_{\widehat{R}}}(\x))^2 d\mu\\  &=&\int_K  \left(\w_2^\top a\left(\W \x \right) - \w_2^\top a\left(\W_{\widehat{R}} \x \right) \right)^2 d\mu \\
&\leq& L \norm{\w_2} \norm{K} \norm{\W-\W_{\widehat{R}}}_F \\
&\leq& L \norm{\w_2} \norm{K} \norm{\mat{R}_w(\W)-\mat{R}_w(\W_{\widehat{R}})}_F \\
&\leq & L \norm{\w_2} \norm{K} \sum_{r=\widehat{R}+1}^R \sigma^2_{r}, 
\end{eqnarray*}
where $\mat{R}_w$ is the reshaping operation in \eqref{eq:nkp-tensor-problem-reshaped}.
Thus, the second term \eqref{eq:second} is bounded by $\epsilon $ if 
$$\sum\limits_{r=\hat R+1}^R\sigma^2_{r} <\epsilon ( L \norm{\w_2} \norm{K})^{-1}.$$ 
Note that the last term \eqref{eq:third} is consequently bounded by the Cauchy-Schwarz inequality.
    \end{proof}

\deepconv*
\begin{proof}
The proof follows a similar proof sketch as in Theorem~\ref{theo:shallow}. Define the $j^{\text{th}}$ convolution layer as
$$h^j(\x) = a(\W^j h^{j-1}(\x)+\w^j_0), $$
where $h^0(\x) = x$, $J$ is the total number of layers and $\W$ of size $d_j\times d_{j-1}$ is a Toeplitz type matrix that transforms a convolution to a matrix multiplication operation. We note that $\tensor W$ is collection of such $\W$'s in a layer.

For a CNN of depth $J$, the hypothesis space is the set of all functions defined by
$$\hat {\mathcal F}^{J}_{\vec{\theta}} = \left\{\sum_{k=1}^{d_J} w_{2k} h^J_k(\x)\right\},$$ 

parametrized by $\vec{\theta} = \{\mat{W}^{j}, \vec{w}_2^j\}_{j=1}^J$

According to \cite[Theorem 1] {Zhou_UniversalDeepCNN_2020}, $\hat{\mathcal F}_\vec{\theta}^J$ is dense in $C(K)$ in $L^\infty$, so it is also dense in $L^2$. In other words, for a given $f \in C(K)$ there exists an approximating convolutional neural network $\hat{f}_\vec{\theta} \in \hat {\mathcal F}_\vec{\theta}^J$, such that
\begin{equation}
    \norm{f - \hat f_\vec{\theta}}_{2, \mu} < \epsilon.
\end{equation}
Building off of this result, it is sufficient to bound a Kronecker convolutional neural network $\hat{f}_{ \vec{\theta}_{\widehat{R}},}$
with a low Kronecker rank $\widehat{R}_j$ in its $j^\text{th}$ hidden layer $h_{{\widehat{R}}_j}^j$
as follows:
\begin{eqnarray*}
&& \norm{\hat{f}_{\vec{\theta}} - \hat{f}_{ \vec{\theta}_{\widehat{R}}
}}_{2, \mu}\\  
&=& \int_K \| \w_2^\top h^J( \W^J  h^{J-1}(\x)) \\
& & \quad\quad\quad - \w_2^\top \hat{h}_{\widehat{R}_J}^J(\W^J_{\widehat{R}_J} h^{J-1}_{\widehat{R}_{J-1}}(\x)) \|_{2}^2 d\mu \\
&  \leq& L \norm{\w_2}_2 \int_K \norm{\W^J-\W_{\widehat{R}_J}^J}_F \\
&& \norm{ h^{J-1}(\W^{J-1}\x) - h_{\widehat{R}_{J-1}}^{J-1}(\W^{J-1}_{\widehat{R}_{J-1}}\x) }_2 d\mu \\
&\leq& L^2 \norm{\w_2}_2 \left( \sum_{r=\widehat{R}_J+1}^{R_J} \sigma^2_{r, J}\right)
\left( \sum_{r=\widehat{ R}_{J-1}+1}^{R_{J-1}} \sigma^2_{r, J-1}\right)\\ 
&&\norm{h^{J-2}(\W^{J-2}\x) - h_{\widehat{R}_{J-2}}^{J-2}(\W^{J-2}_{\widehat{R}_{J-2}}\x) }_2, 
\end{eqnarray*}


expanding on inner layers gives
$$\leq L^{J}\norm{\w_2} \norm{K} \prod_{j=1}^J \left(\sum_{r={\widehat{R}}_j +1}^{R_j}\sigma^2_{r, j}\right),$$ and the therefore the low rank network is bounded by $\epsilon$ given  $$\prod_{j=1}^J \left(\sum_{r=\widehat{R}_j +1}^{R_j}\sigma^2_{r, j}\right) < \epsilon(L^J\norm{\w_2} \norm{K})^{-1}.$$
\end{proof}

\subsection{Reconstruction Error of ResNet18}

\begin{figure}[h]
         \centering
         \includegraphics[width=0.9\columnwidth]{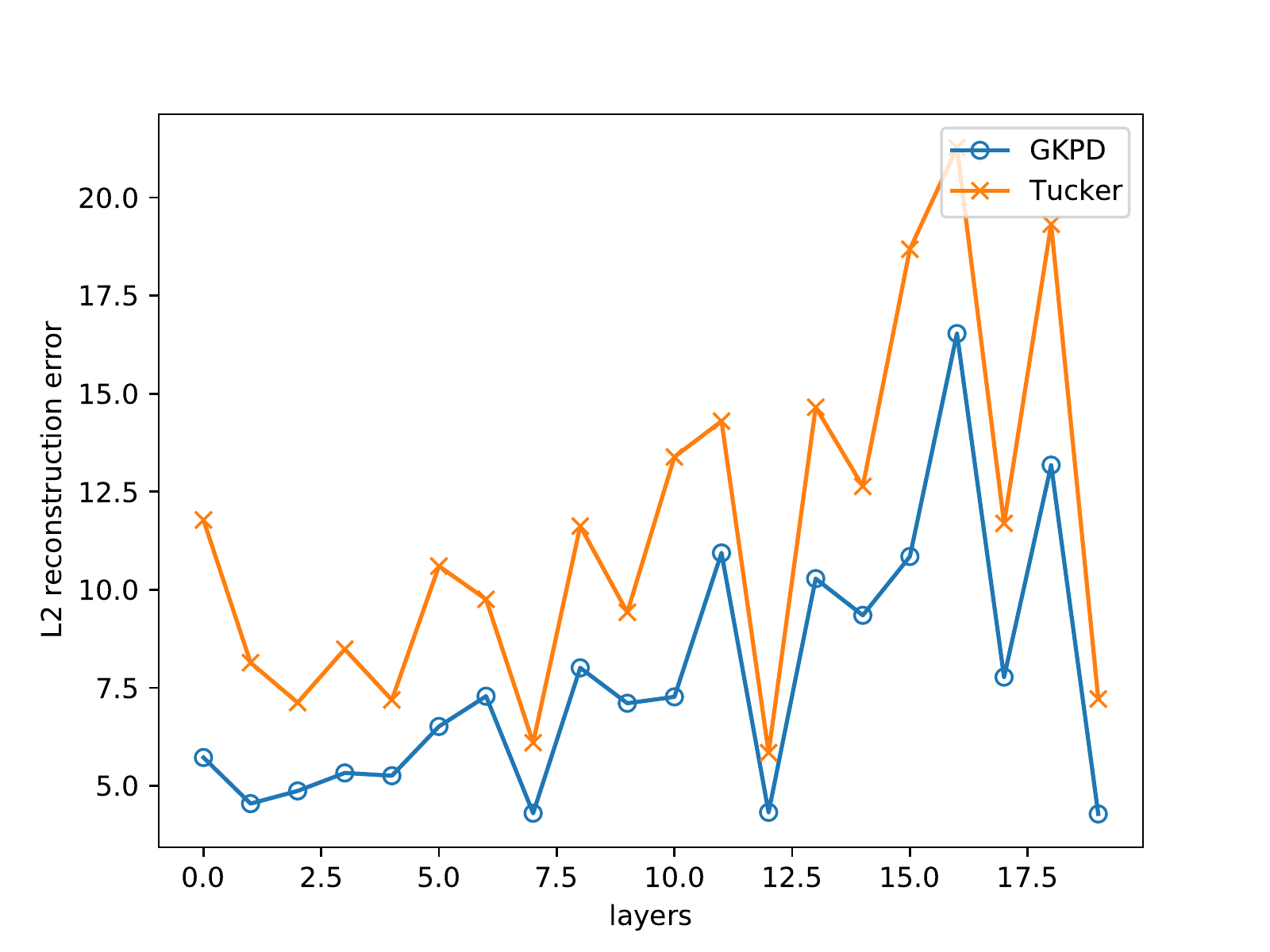}
         \caption{Reconstruction error between convolution tensors in a pretrained ResNet18 model and compressed representations at a $4\times$ compression rate. GKPD always yields a lower reconstruction error than Tucker decomposition}
         \label{fig:reconstruction_loss}
\end{figure}

We further study GKPD by plotting the $L_2$ reconstruction errors achieved when compressing a ResNet18 model that is pretrained on ImageNet. We obseve in Figure~\ref{fig:reconstruction_loss} that GKPD generally achieves a lower reconstruction error in comparison with Tucker decomposition.

\hfill


\end{document}